\documentclass{article}

\usepackage{arxiv}

\usepackage[utf8]{inputenc} % allow utf-8 input
\usepackage[T1]{fontenc}    % use 8-bit T1 fonts
\usepackage{hyperref}       % hyperlinks
\usepackage{url}            % simple URL typesetting
\usepackage{booktabs}       % professional-quality tables
\usepackage{amsfonts}       % blackboard math symbols
\usepackage{nicefrac}       % compact symbols for 1/2, etc.
\usepackage{microtype}      % microtypography

\usepackage{amsmath, amssymb, amsthm}
\usepackage{algorithm, algorithmic}

\usepackage{mathtools}

\usepackage{caption}
\usepackage{subcaption}
\usepackage{color}
\usepackage{arydshln}

\usepackage{algorithm}
\usepackage{algorithmic}
\usepackage{amsmath,amssymb}
\usepackage{tikz}
\usetikzlibrary{positioning, arrows.meta}
\allowdisplaybreaks[4]

\title{Enhancing binary classification: A new stacking method via leveraging computational geometry}

\author{
    Wei~Wu\thanks{Co-corresponding author. email: \texttt{goi@shizuoka.ac.jp}.}\\
    Shizuoka University\\
    \And
    Liang~Tang\thanks{Co-corresponding author. email: \texttt{tangericliang@dlmu.edu.cn}.}\\
    Dalian Maritime University\\
    \And
    Zhongjie~Zhao\\
    Dalian Maritime University\\
    \And
    Chung-Piaw~Teo\\
    National University of Singapore\\
}

\begin{document}
\maketitle

\begin{abstract}
Stacking, a potent ensemble learning method, leverages a meta-model to harness the strengths of multiple base models, thereby enhancing prediction accuracy. Traditional stacking techniques typically utilize established learning models, such as logistic regression, as the meta-model. This paper introduces a novel approach that integrates computational geometry techniques, specifically solving the maximum weighted rectangle problem, to develop a new meta-model for binary classification. Our method is evaluated on multiple open datasets, with statistical analysis showing its stability and demonstrating improvements in accuracy compared to current state-of-the-art stacking methods with out-of-fold predictions. This new stacking method also boasts two significant advantages: enhanced interpretability and the elimination of hyperparameter tuning for the meta-model, thus increasing its practicality. These merits make our method highly applicable not only in stacking ensemble learning but also in various real-world applications, such as hospital health evaluation scoring and bank credit scoring systems, offering a fresh evaluation perspective.
\end{abstract}

% keywords can be removed

\section{Introduction}
Binary classification is a fundamental task in machine learning and data science, with applications spanning numerous domains, including spam detection, medical diagnostics, image recognition, credit scoring. The goal is to predict a binary outcome—typically labeled as 0 or 1—based on a set of input features. Various machine learning algorithms, such as logistic regression (LR), $k$-nearest neighbors ($k$NN), support vector machines (SVM), and neural network (NN), are commonly employed for binary classification tasks.
These algorithms can be mainly divided into two categories: those with interpretability, which are convenient for analysis and control (e.g., LR); and those without interpretability but with potentially good classification performance (e.g., NN).

Ensemble learning, a powerful technique in predictive modeling, has gained widespread recognition for its ability to improve model performance by combining the strengths of multiple learning algorithms~\cite{dong2020survey}. Among ensemble methods, stacking stands out by integrating the predictions of diverse base models (different learning algorithms) through a meta-model, resulting in enhanced prediction accuracy compared to only using the best base model~\cite{dvzeroski2004combining}. Stacking has demonstrated significant applications in classification problems such as network intrusion detection~\cite{rajagopal2020stacking,thockchom2023novel}, 
cancer type classification~\cite{mohammed2021stacking},
credit lending~\cite{muslim2023new},
and protein-protein binding affinity prediction~\cite{xiang2022persistent}.

Many researchers have been working on configuring the components of stacking, selecting adaptive base models and meta-model from a set of candidates by meta-heuristics~\cite{gupta2014optimization}, including genetic algorithm~\cite{ordonez2008genetic}, data envelopment analysis~\cite{zhu2010hybrid} and ant colony optimization~\cite{chen2014applying}. However, the design of the meta-model itself in stacking has largely been limited to conventional machine learning algorithms.

In this paper, we propose a novel stacking ensemble learning framework that leverages the maximum weighted rectangle problem (MWRP), a well-known problem in computational geometry, to construct an effective meta-model. The essence of the MWRP-based stacking lies in transforming the one-dimensional probability outputs of multiple base models for individual sample points into a multi-dimensional space. This transformation enriches the representation of the sample points with more high-dimensional information, which in turn allows the method to identify the optimal combination of base models and corresponding thresholds on selected models. The goal is to construct a rectangle in this high-dimensional space that maximizes the expected number of correctly classified sample points within it.

The MWRP-based stacking framework not only selects base models but also provides valuable insights into the selected models, thereby maintaining high interpretability.
Through extensive computational experiments across various datasets, we demonstrate the efficacy of our approach in enhancing predictive accuracy compared to traditional stacking methods.
Furthermore, we explore potential extensions of the MWRP-based technique, particularly its application as an interpretable binary classifier and its capability to effectively handle imbalanced datasets.

\section{Binary Classification by Stacking Ensemble Learning}
In this section, we discuss the background of supervised binary classification and the fundamental principles of the stacking ensemble learning framework.
\subsection{Binary Classification}
Let $\mathcal{X}=\mathbb{R}^m$ denote the $m$-dimensional feature space, and $\mathcal{Y}_{\mathrm{B}}=\{0, 1\}$ denote the label space.
Given a dataset $D =((x_1,y_1),(x_2,y_2),\ldots,(x_n,y_n))$ with pairs of training samples $x_i\in \mathcal{X}$ and $y_i\in \mathcal{Y}_{\mathrm{B}}$ for $i=1,2,\ldots,n$, binary classification aims to
predict the output $y \in \mathcal{Y}_{\mathrm{B}}$ of a new input $x \in \mathcal{X}$.

Binary classification is a fundamental task in data science and machine learning, with applications that extend across numerous domains, including spam detection, medical diagnosis, image analysis, quality assurance, credit assessment, and marketing.
Each of these applications presents unique challenges, such as varying data distributions, noise levels, and feature correlations, which influence the choice of the appropriate machine learning method.

A variety of machine learning algorithms are employed for binary classification.
% , including logistic regression, $k$-nearest neighbors ($k$NN), support vector machines (SVM), decision trees, neural networks, and gradient boosting. 
The selection of a particular method is typically influenced by the characteristics of the dataset and the specific requirements of the problem at hand.
For example, logistic regression is often preferred for its simplicity and ease of interpretation, especially when the relationship between the features and the log-odds of the target variable is approximately linear.
In contrast, methods like SVM or gradient boosting might be preferred for their ability to handle more complex, non-linear relationships and high-dimensional data.

\subsection{Stacking Ensemble Learning}
Ensemble learning, in particular, enhances predictive performance by integrating the outputs of several models, thereby surpassing the effectiveness of any single model alone.
Techniques such as, bagging, boosting and stacking, each capture different aspects of the data, contributing to improved prediction accuracy.

Among these, stacking is an ensemble learning technique that integrates predictions from various base models, often referred to as first-level models, and uses these predictions to train a meta-model, which then produces the final prediction.
The strength of stacking lies in its ability to synthesize the complementary strengths of diverse base models, leading to more robust and accurate predictions.

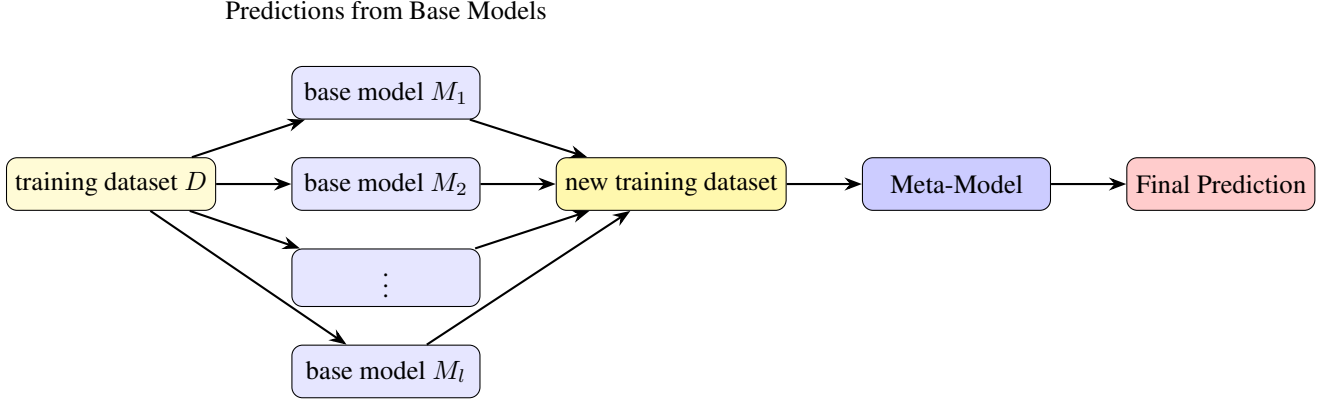
\begin{figure*}[ht]
\begin{tikzpicture}[
    scale=0.1,
    base/.style={rectangle, draw, rounded corners, minimum height=0.7cm, minimum width=2.5cm, text centered},
    arrow/.style={-Stealth, thick},
    ]

    % Nodes for Base Models
    \node[base, fill=blue!10] (model1) {base model $M_1$};
    \node[base, fill=blue!10, below=0.5cm of model1] (model2) {base model $M_2$};
    \node[base, fill=blue!10, below=0.5cm of model2] (model3) {$\vdots$};
    \node[base, fill=blue!10, below=0.5cm of model3] (modelm) {base model $M_l$};

    % Node for Dataset
    \node[base, fill=yellow!20, left=1cm of model2] (dataset) {training dataset $D$};

    % Node for New Dataset
    \node[base, fill=yellow!40, right=1cm of model2] (ndataset) {new training dataset};

    % Node for Meta-Model
    \node[base, fill=blue!20, right=1cm of ndataset] (metamodel) {Meta-Model};

    % Node for Final Prediction
    \node[base, fill=red!20, right=1cm of metamodel] (finalprediction) {Final Prediction};

    % Arrows from dataset to base models
    \draw[arrow] (dataset) -- (model1);
    \draw[arrow] (dataset) -- (model2);
    \draw[arrow] (dataset) -- (model3);
    \draw[arrow] (dataset) -- (modelm);

    % Arrows from base models to meta-model
    \draw[arrow] (model1) -- (ndataset);
    \draw[arrow] (model2) -- (ndataset);
    \draw[arrow] (model3) -- (ndataset);
    \draw[arrow] (modelm) -- (ndataset);

    % Arrow from meta-model to final prediction
    \draw[arrow] (ndataset) -- (metamodel);

    % Arrow from meta-model to final prediction
    \draw[arrow] (metamodel) -- (finalprediction);

    % Labels
    \node[above=0.5cm of model1] {Predictions from Base Models};

\end{tikzpicture}
\caption{Framework of stacking ensemble learning with $l$ base models.}
\label{fig:stacking}
\end{figure*}

In the stacking process, multiple base models are first trained independently on the same training dataset. The outputs of these base models are then combined to form a new dataset, while the original labels are retained as the labels for this new training data. This new dataset is used to train the meta-model, also known as the second-level model. The meta-model synthesizes the predictions from the base models to produce the final prediction.
The framework of stacking is shown in \figurename~\ref{fig:stacking}.

% By harnessing the strengths of various base models while mitigating their individual weaknesses, stacking can significantly enhance prediction accuracy. This approach is especially advantageous in scenarios involving diverse data sources and complex problems, owing to its adaptability and robust performance.
% As a result, stacking is widely used in machine learning competitions and real-world applications where top-tier performance is essential. Implementing this method allows for the construction of more accurate predictive models, thereby improving the quality of data-driven decision-making.

The choice of the meta-model plays a crucial role in the success of the stacking ensemble. Although stacking has been widely applied in practical scenarios, the design of the meta-model has typically been confined to only selecting basic learning methods, such as LR, SVM, and NN. To the best of our knowledge, no studies have explored the use of computational geometry in designing a meta-model, which represents a significant gap that this paper aims to address.

\subsection{Out-of-Fold Predictions}
In most cases, stacking can yield better results than individual base models.
However, if some base models are overfitting, the meta-model may become dominated by these overfitting models, leading to reduced generalization performance.
To mitigate this issue, out-of-fold predictions are used as an effective technique. This approach is typically integrated into the $k$-fold cross-validation process to prepare unbiased input data for the meta-model.

In the $k$-fold cross-validation process, the dataset $D$ is divided into $k$ subsets or folds, $F_1, F_2, \ldots, F_k$. For each fold $F_i$, a model is trained on the remaining $k-1$ folds and validated on the left-out fold $F_i$. We use $Y^{(j)}_i$ to denote the predictions made by model $M_j$ on fold $F_i$. These predictions $Y^{(j)}_i$ are referred to as out-of-fold predictions because they are generated from data not seen during training.
This process is repeated $k$ times, ensuring that every sample in the dataset has a prediction generated when it was excluded from training.
Combining $Y^{(j)}_1, Y^{(j)}_2, \ldots, Y^{(j)}_k$ forms the complete set of predictions for model $M_j$, which can be used as an input feature for a meta-model in stacking.

The pseudocode for stacking ensemble learning with out-of-fold predictions with $l$ base models is shown in Algorithm~\ref{algo:oof}.

\begin{algorithm}
\caption{Stacking ensemble learning with out-of-fold predictions.}\label{algo:oof}
\begin{algorithmic}[1]
\STATE \textbf{Input:} Dataset $D$, number of folds $k$, set of base models $\{M_1, M_2, \ldots, M_l\}$.
\STATE Split $D$ into $k$ folds: $F_1, F_2, \ldots, F_k$.
\STATE Initialize an empty matrix $Y_\mathrm{oof} \in \mathcal{Y}_{\mathrm{B}}^{n \times l}$ for storing out-of-fold predictions.\label{step:prepare}
\FOR{each model $M_j$ in $\{M_1, M_2, \ldots, M_l\}$}
    \FOR{each fold $F_i$ in $\{F_1, F_2, \ldots, F_k\}$}
        \STATE $\mathcal{D}_{\text{train}} \gets D \setminus F_i$. %\COMMENT{Use all data except the $i$-th fold for training}
        \STATE $\mathcal{D}_{\text{test}} \gets F_i$. %\COMMENT{Use the $i$-th fold as the validation set}
        \STATE Train model $M_j$ on $\mathcal{D}_{\text{train}}$.
        \STATE Use $M_j$ to obtain binary predictions $Y^{(j)}_i$ on $\mathcal{D}_{\text{test}}$. \label{step:train}%\COMMENT{Generate out-of-fold predictions for $F_i$}
    \ENDFOR
    \STATE Combining $Y^{(j)}_1, Y^{(j)}_2, \ldots, Y^{(j)}_k$ to form the complete predictions $Y^{(j)}$ of model $M_j$.
    \STATE Store $Y^{(j)}$ in the corresponding column of $Y_\mathrm{oof}$.
\ENDFOR
\STATE Train meta-model $M_{\text{meta}}$ using $Y_\mathrm{oof}$ as input features.
\STATE \textbf{Output:} Trained meta-model $M_{\text{meta}}$.
\end{algorithmic}
\end{algorithm}

% \color{blue}The key advantages of out-of-fold predictions include offering an unbiased estimate of model performance, reducing the risk of overfitting by ensuring the meta-model does not learn from training data predictions, and improving the model's generalization capability.
% This technique is crucial for building robust models that perform effectively on unseen data, making it an invaluable tool for both research and real-world applications.\color{black}

\section{Stacking Ensemble Learning via Computational Geometry}
In this section, we begin by introducing a computational geometry problem, the maximum weighted rectangle problem, and then we propose a new stacking ensemble learning approach that centers on solving this problem.
\subsection{Maximum Weighted Rectangle Problem}
We are given $\tilde{n}$ weighted points in an $\tilde{m}$-dimensional space, where each point $i$ has a weight $w_i$ that can be either positive or negative. Let $a_{ij}$ represent the coordinate of point $i$ in dimension $j$. The maximum weighted rectangle problem (MWRP) aims to find an $\tilde{m}$-dimensional axis-aligned rectangle (with each face perpendicular to the dimensional axes) that maximizes the total weight of the points contained within the box.

To formally explain the MWRP, we propose a mathematical model. Let continuous variables $\alpha_j^\mathrm{lb}$ and $\alpha_j^\mathrm{ub}$ represent the lower and upper bounds of the rectangle in dimension $j$, respectively. For a point $i$, let auxiliary binary variable $\beta_i$ indicate whether point $i$ is inside the rectangle (1 if inside, 0 if outside), and let auxiliary binary variables $\gamma_{ij}^\mathrm{lb}$ and $\gamma_{ij}^\mathrm{ub}$ indicate whether the lower and upper bounds of the rectangle are satisfied in dimension $j$, respectively. By introducing large constants $b_j=1+\max_{i=1}^{\tilde n}a_{ij}-\min_{i=1}^{\tilde n} a_{ij}$, the mixed integer programming (MIP) model for the MWRP can be formulated as follows:
\begin{align}
\nonumber&\text{maximize}\\
\label{mwr:obj}&\quad\sum_{j=1}^{\tilde{n}} w_i \beta_i \\
\nonumber&\text{subject to} \\
\nonumber&\quad\alpha_j^\mathrm{lb} - b_j(1 - \beta_i) \leq a_{ij} \leq \alpha_j^\mathrm{ub} + b_j(1 - \beta_i) \\
\label{mwr:beta2inq}
&\hspace{34mm}\forall i = 1, \dots, \tilde{n}; \forall j = 1, \dots, \tilde{m} \\
\label{mwr:inq2beta-1}
&\quad b_j\gamma_{ij}^\mathrm{lb} > a_{ij}-\alpha_j^\mathrm{lb}\hspace{5mm}\forall i = 1, \dots, \tilde{n}; \forall j = 1, \dots, \tilde{m} \\
\label{mwr:inq2beta-2}&\quad b_j\gamma_{ij}^\mathrm{ub} > \alpha_j^\mathrm{ub} - a_{ij}\hspace{3mm}\forall i = 1, \dots, \tilde{n}; \forall j = 1, \dots, \tilde{m} \\
\nonumber&\quad 2\tilde{m}\beta_i \le \sum_{j=1}^{\tilde{m}}(\gamma_{ij}^\mathrm{lb}+\gamma_{ij}^\mathrm{ub}) \le \beta_i + 2\tilde{m} - 1\\
\label{mwr:inq2beta-3}
& \hspace{56mm} \forall i = 1, \dots, \tilde{n}\\
&\quad \gamma_{ij}^\mathrm{lb},\gamma_{ij}^\mathrm{ub} \in \{0, 1\}\hspace{6mm}\forall i = 1, \dots, \tilde{n}; \forall j = 1, \dots, \tilde{m} \\
&\quad \beta_i \in \{0, 1\}, \hspace{34mm} \forall i = 1, \dots, \tilde{n}.
\end{align}
The objective function~\eqref{mwr:obj} maximizes the total weight of the points within the rectangle.
Constraints~\eqref{mwr:beta2inq} ensure that for each point $i$ to be inside the rectangle (i.e., $\beta_i = 1$), the following constraints must hold:
\begin{align}\label{cond}
&\alpha_j^\mathrm{lb} \leq a_{ij} \leq \alpha_j^\mathrm{ub},&\forall j = 1, \dots, \tilde{m}.
\end{align}
Constraints~\eqref{mwr:inq2beta-1}--\eqref{mwr:inq2beta-3} indicate that if condition \eqref{cond} holds for point $i$, then $\beta_i$ must take the value 1.
A solution of the MWRP can be represented by the combination of $\boldsymbol{\alpha}^\mathrm{lb}=(\alpha_1^\mathrm{lb},\alpha_2^\mathrm{lb},\ldots,\alpha_{\tilde{m}}^\mathrm{lb})$ and $\boldsymbol{\alpha}^\mathrm{ub}=(\alpha_1^\mathrm{ub},\alpha_2^\mathrm{ub},\ldots,\alpha_{\tilde{m}}^\mathrm{ub})$.

The MWRP has been studied for a long time in the field of computational geometry.
However, to the best of our knowledge, only exact algorithms with exponential computation time of $\Omega(\tilde{n}^{\tilde{m}})$ have been proposed \cite{backurs2016tight}. Due to the computational complexity, there has been less consideration of real-world applications, particularly in machine learning.

\subsection{Stacking by Solving the MWRP}
Before demonstrating how the MWRP can be utilized to design a new meta-model in stacking, we note that in Algorithm~\ref{algo:oof}, probability predictions from the base models can be used instead of binary predictions to train the meta-model $M_{\text{meta}}$. More specifically, in Step~\ref{step:prepare}, $Y_\mathrm{oof} \in \mathcal{Y}_{\mathrm{B}}^{n \times l}$ is replaced with $\bar{Y}_\mathrm{oof} \in \mathcal{Y}_{\mathrm{R}}^{n \times l}$, where $\mathcal{Y}_{\mathrm{R}} = [0,1]$ represents the predicted probability of label 1.
Then, in Step~\ref{step:train}, we use base model $M_j$ to obtain probability predictions to form $\bar{Y}_\mathrm{oof}$.

For the meta-model dataset $\bar{Y}_\mathrm{oof}$, let $p_{ij}$ denote the probability prediction of base model $M_j$ for sample $i$. This dataset can then be transformed into an instance of the MWRP, where each sample $i$ is represented as a point in an $l$-dimensional space, with the coordinate in dimension $j$ corresponding to $p_{ij}$, that is, $a_{ij}=p_{ij}$. Additionally, the weight $w_i$ for each sample $i$ in the MWRP is set as follows:
\begin{align}\label{0mwr}
w_i = \begin{cases} 
1 & \text{if label $y_i$ is 0}, \\
-1 & \text{if label $y_i$ is 1}.
\end{cases}
\end{align}
% We refer to the MWRP translated by with~\eqref{0mwr} as 0-MWRP.

\begin{figure}[t]
\centering
\includegraphics[width=0.7\columnwidth]{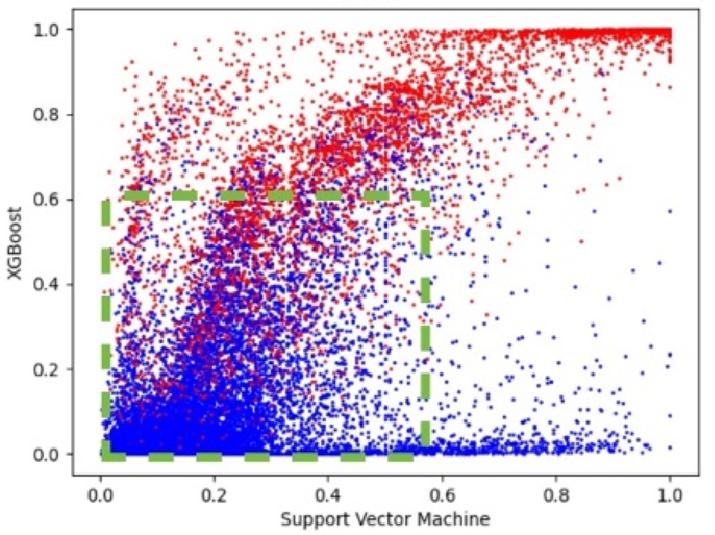}
\caption{MWRP in meta-model training.}
\label{fig:mwrp}
\end{figure}

By solving the corresponding MWRP, the obtained rectangle represents the trained meta-model.
\figurename~\ref{fig:mwrp} illustrates an example using the MWRP, where there are only two base models.
The blue points represent samples with a label of 0 and are assigned a weight of 1, while the red points represent samples with a label of 1 and are assigned a weight of -1. The green rectangle indicates the solution of the MWRP.

With the trained model (the rectangle obtained by solving the MWRP), when a new test sample is provided, it is classified as 0 if it falls inside the rectangle and as 1 if it falls outside.
If the upper and lower bounds of the rectangle in any dimension $j$ are both set to 0 and 1, respectively, it indicates that base model $M_j$ is not being used. Otherwise, the values of the bounds imply the importance of base model $M_j$ on classifying samples.
The MWRP-based meta-model can thus represent both the selection of base models and the extent to which each selected model contributes to the final decision, offering high interpretability and low complexity.

Alternatively, we can also consider an MWRP where the weight $w_i$ is defined as follows:
\begin{align}\label{1mwr}
w_i = \begin{cases} 
1 & \text{if label $y_i$ is 1}, \\
-1 & \text{if label $y_i$ is 0}.
\end{cases}
\end{align}
In this case, a test sample is classified as 1 if it falls inside the rectangle and as 0 if it falls outside. We refer to the meta-models using the weight settings in \eqref{0mwr} and \eqref{1mwr} as the 0-MWRP model and 1-MWRP model, respectively.

For the transformed MWRPs, because each base model is relatively accurate, a point corresponding to a label-0 (resp., label-1) sample is more likely to be close to $\boldsymbol{0}_m$ (resp., $\boldsymbol{1}_m$).
Thus, we can reduce the computational effort by fixing $\boldsymbol{\alpha}^\mathrm{lb} = \boldsymbol{0}_m$ for the 0-MWRP and $\boldsymbol{\alpha}^\mathrm{ub} = \boldsymbol{1}_m$ for the 1-MWRP. Consequently, a solution to the 0-MWRP (resp. 1-MWRP) can be simply represented by $\boldsymbol{\alpha}^\mathrm{ub}$ (resp. $\boldsymbol{\alpha}^\mathrm{lb}$).

When both the 0-MWRP and 1-MWRP are solved, we propose a new strategy, which we call mixed-MWRP, to predict the label of a sample by considering both solutions (rectangles). 
Let $\boldsymbol{\bar{\alpha}}^\mathrm{ub}$ and $\boldsymbol{\bar{\alpha}}^\mathrm{lb}$ be the solutions obtained from the 0-MWRP and 1-MWRP, respectively.
For a sample $i$, the following four cases exist:
\begin{enumerate}
\item\label{case:inout} $i$ is inside the rectangle obtained by 0-MWRP and outside the rectangle obtained by 1-MWRP;
\item\label{case:outin} $i$ is outside the rectangle obtained by 0-MWRP and inside the rectangle obtained by 1-MWRP;
\item\label{case:outout} $i$ is outside the rectangles obtained by both 0-MWRP and 1-MWRP;
\item\label{case:inin} $i$ is inside the rectangles obtained by both 0-MWRP and 1-MWRP.
\end{enumerate}
In Cases~\ref{case:inout} and \ref{case:outin}, there is no conflict between the two solutions, so the label prediction is 0 for Case~\ref{case:inout} and 1 for Case~\ref{case:outin}.
% For Case~\ref{case:outout}, we set the label prediction to 0 if $d(i, \boldsymbol{\bar{\alpha}}^\mathrm{ub}) \leq d(i, \boldsymbol{\bar{\alpha}}^\mathrm{lb})$ and 1 otherwise, where
% \begin{align*}
% &d(i, \boldsymbol{\bar{\alpha}}^\mathrm{ub}) = \max_{j=1}^{l} \left\{p_{ij}-\bar{\alpha}_j^\mathrm{ub}\right\},\\
% &d(i, \boldsymbol{\bar{\alpha}}^\mathrm{lb}) = \max_{j=1}^{l} \left\{\bar{\alpha}_j^\mathrm{lb}-p_{ij}\right\}.
% \end{align*}
For Case~\ref{case:outout}, we compute the Euclidean distances between $i$ and the two rectangles:
\begin{align*}
&d(i, \boldsymbol{\bar{\alpha}}^\mathrm{ub}) = \sqrt{\sum_{j=1}^{l} \left(\max\left\{0, p_{ij} - \bar{\alpha}_j^\mathrm{ub}\right\}\right)^2},\\
&d(i, \boldsymbol{\bar{\alpha}}^\mathrm{lb}) = \sqrt{\sum_{j=1}^{l} \left(\max\left\{0, \bar{\alpha}_j^\mathrm{lb} - p_{ij}\right\}\right)^2}.
\end{align*}
Then, we assign the label according to the rectangle with the shorter distance; that is, 0 if $d(i, \boldsymbol{\bar{\alpha}}^\mathrm{ub}) \leq d(i, \boldsymbol{\bar{\alpha}}^\mathrm{lb})$, and 1 otherwise.
For Case~\ref{case:inin}, we set the label prediction to 0 if $d'(i, \boldsymbol{\bar{\alpha}}^\mathrm{ub}) \geq d'(i, \boldsymbol{\bar{\alpha}}^\mathrm{lb})$ and 1 otherwise, where
\begin{align*}
&d'(i, \boldsymbol{\bar{\alpha}}^\mathrm{ub}) = \min_{j=1}^{l} \left\{\bar{\alpha}_j^\mathrm{ub} - p_{ij}\right\},\\
&d'(i, \boldsymbol{\bar{\alpha}}^\mathrm{lb}) = \min_{j=1}^{l} \left\{p_{ij} - \bar{\alpha}_j^\mathrm{lb}\right\}.
\end{align*}

\subsection{Metaheuristic Design}
Although the MWRPs, 0-MWRP and 1-MWRP, can be exactly solved by the $\Omega(n^m)$-time algorithm \cite{backurs2016tight} or the MIP solver using the MIP model~\eqref{mwr:obj}--\eqref{mwr:inq2beta-3}, preliminary experimental results showed that obtaining an optimal solution in a reasonable time is not possible for a large-scale dataset with multiple base models.

In this paper, we propose an iterated local search (ILS) algorithm that can obtain a good solution even when the training time is limited.
ILS is a metaheuristic optimization algorithm used to find a good (often near-optimal) solution to difficult combinatorial problems.
It is based on the idea of iteratively improving a solution by applying local search techniques combined with perturbations to escape local optima~\cite{lourencco2019iterated}.

We introduce the ILS for the 0-MWRP. (The one for the 1-MWRP can be implemented similarly.)
First, we generate an initial solution $\boldsymbol{\alpha}^\mathrm{ub}$ using a greedy algorithm that sequentially fixes each $\alpha_j^\mathrm{ub}$ to maximize the current total weight.
Based on this initial solution $\boldsymbol{\alpha}^\mathrm{ub}$, a local search algorithm starts and attempts to find an improved solution $\boldsymbol{\tilde{\alpha}}^\mathrm{ub}$ among neighboring solutions (the neighborhood) by applying a two-change operation in which the upper bounds of the rectangle can be adjusted in at most two dimensions.
If such an improved solution $\boldsymbol{\tilde{\alpha}}^\mathrm{ub}$ is found, the local search continues from $\boldsymbol{\tilde{\alpha}}^\mathrm{ub}$, and this process is repeated until a local optimal solution is found (i.e., no better solution exists in the neighborhood).
Note that if the number of dimensions $l$ (the number of base models) is only 2, the local search can provide an optimal solution.

After obtaining a local optimal solution, the ILS algorithm applies a ``kick'' to facilitate diversification, allowing the local search to find a better solution.
A kick modifies the current solution and should be strong enough to escape local minima, but weak enough to retain the good characteristics of the current locally optimal solution.
Weak kicks can lead to faster local search, as good starting points can quickly provide locally optimal solutions.
To design a kick for the 0-MWRP, we randomly change the upper bounds of the rectangle in $q$ dimensions from the local optimal solution. 

In this paper, we use the following simple idea for adjusting the value of $q$.
We first set $q$ to an initial value of 3, because $q=2$ will not be able to get rid of the obtained local optimal solution.
If the current local search does not find a better solution than the last local search, we increase $q$ by one. Otherwise, if an improved solution is found, we reset $q$ to 3 again.

In the ILS for the \mbox{0-MWRP}, we always apply a kick to the incumbent solution, which is the best local optimal solution found so far.

\section{Computational Experiments}
We tested our approach on multiple datasets.
In this section, we describe the computational results and present some observations based on them.

\subsection{Datasets}
In our computational experiments, we used available binary classification datasets from the \texttt{sklearn.datasets} and the UCI Machine Learning Repository:
\begin{itemize}
    \item \textbf{Adult Income Dataset}\footnote{\url{https://archive.ics.uci.edu/ml/machine-learning-databases/adult/adult.data}} (Income): The Adult dataset, often used for classification tasks, contains census data that is used to predict whether an individual's income exceeds \$50,000 per year based on demographic information.
    
    \item \textbf{Bank Marketing Dataset}\footnote{\url{https://archive.ics.uci.edu/ml/machine-learning-databases/00222/bank-additional.zip}} (Bank): This dataset contains data related to direct marketing campaigns of a Portuguese banking institution. The goal is to predict whether a client will subscribe to a term deposit.
    
    \item \textbf{Labeled Faces in the Wild Pairs Dataset}\footnote{sklearn.datasets.fetch\_lfw\_pairs} (LFW): This dataset contains pairs of images of faces labeled as either the same person or different people. It is used for tasks related to face verification and contains 2,200 pairs.
    
    \item \textbf{Banknote Authentication Dataset}\footnote{\url{https://archive.ics.uci.edu/ml/machine-learning-databases/00267/data_banknote_authentication.txt}} (Banknote): This dataset contains features extracted from images of genuine and forged banknotes. The goal is to classify the banknotes as either authentic or forged.
    
    \item \textbf{German Credit Dataset}\footnote{\url{https://archive.ics.uci.edu/ml/machine-learning-databases/statlog/german/german.data}} (Credit): This dataset is used for credit risk classification, where the task is to predict the creditworthiness of individuals based on their personal and financial information.
    
    \item \textbf{Breast Cancer Wisconsin (Diagnostic) Dataset}\footnote{sklearn.datasets.load\_breast\_cancer} (Cancer): This dataset consists of features computed from a digitized image of a fine needle aspirate of a breast mass. The goal is to classify the tumor as malignant or benign.
    
    \item \textbf{Ionosphere Dataset}\footnote{\url{https://archive.ics.uci.edu/ml/machine-learning-databases/ionosphere/ionosphere.data}} (Ionosphere): This dataset contains radar data collected by a system in Goose Bay, Labrador. The goal is to classify the returns from radar signals as either good or bad.
    
    \item \textbf{Sonar Dataset}\footnote{\url{https://archive.ics.uci.edu/ml/machine-learning-databases/undocumented/connectionist-bench/sonar/sonar.all-data}} (Sonar): This dataset contains data obtained from a sonar signal. The goal is to classify objects as either a rock or a metal cylinder based on the returned signal.
\end{itemize}

These datasets were chosen to represent a wide range of domains and challenges, allowing us to thoroughly evaluate the performance and generalizability of the proposed methods across different scenarios.
The summary of these datasets is shown in \tablename~\ref{tbl:dataset}, where columns `\#samples' and `\#features' show the number of samples and features for each dataset.

\begin{table}[ht]
\centering \fontsize{9pt}{11pt}\selectfont
    \caption{Summary of datasets used in experiments.}\label{tbl:dataset}
    \begin{tabular}{lrrr}
        \hline
        dataset && \#samples & \#features \\ \hline
        Income && 48,842 & 14 \\ 
        Bank && 41,188 & 20 \\ 
        LFW && 2,200 & 5,828  \\ 
        Banknote && 1,372 & 4 \\ 
        Credit && 1,000 & 61 \\ 
        Cancer && 569 & 30 \\ 
        Ionosphere && 351 & 34 \\ 
        Sonar && 208 & 60 \\ \hline
    \end{tabular}
\end{table}

\begin{table*}[t]
\centering \fontsize{9pt}{11pt}\selectfont
    \centering
    \caption{Comparison results in percentage accuracy with base models.}\label{tbl:accuracy-single}
    \begin{tabular}{lrrrrrrrrrrrr}
    \hline
    && \multicolumn{6}{c}{Single base model} && \multicolumn{3}{c}{Stacking with MWRP}                            \\ \cline{3-8} \cline{10-12} 
    dataset && \multicolumn{1}{c}{RF} & \multicolumn{1}{c}{SVM} & \multicolumn{1}{c}{$k$NN} & \multicolumn{1}{c}{LR} & \multicolumn{1}{c}{XGB} & \multicolumn{1}{c}{FNN} && \multicolumn{1}{c}{0-MWRP} & \multicolumn{1}{c}{1-MWRP} & \multicolumn{1}{c}{mix-MWRP}\\ \hline
Income	&&	84.35\%	&	80.64\%	&	82.83\%	&	81.53\%	&	86.84\%	&	85.08\%	&&	86.46\%	&	86.41\%	&	86.44\%	\\
Bank	&&	91.43\%	&	90.23\%	&	90.12\%	&	90.92\%	&	91.33\%	&	91.08\%	&&	91.36\%	&	91.78\%	&	91.39\%	\\
LFW	&&	63.64\%	&	54.32\%	&	56.36\%	&	56.14\%	&	61.59\%	&	62.50\%	&&	62.27\%	&	63.18\%	&	64.09\%	\\
Banknote	&&	99.64\%	&	98.18\%	&	100.00\%	&	97.45\%	&	99.27\%	&	97.82\%	&&	99.64\%	&	99.64\%	&	100.00\%	\\
Credit	&&	77.00\%	&	76.50\%	&	72.50\%	&	75.00\%	&	74.50\%	&	75.00\%	&&	77.00\%	&	79.50\%	&	77.50\%	\\
Cancer	&&	95.61\%	&	96.49\%	&	95.61\%	&	97.37\%	&	96.49\%	&	95.61\%	&&	96.49\%	&	97.37\%	&	98.25\%	\\
Ionosphere	&&	92.96\%	&	91.55\%	&	84.51\%	&	91.55\%	&	91.55\%	&	90.14\%	&&	90.14\%	&	91.55\%	&	92.96\%	\\
Sonar	&&	76.19\%	&	66.67\%	&	69.05\%	&	71.43\%	&	80.95\%	&	73.81\%	&&	83.33\%	&	90.48\%	&	90.48\%	\\\hline
Average	&&	85.10\%	&	81.82\%	&	81.37\%	&	82.67\%	&	85.32\%	&	83.88\%	&&	85.84\%	&	87.49\%	&	87.64\%	\\\hline
    \end{tabular}
\end{table*}
\subsection{Computational Environment and Implementation Details}
All experiments were carried out on a MacBook Pro with an M1 Max CPU and 64 GB of memory. The ILS for solving the MWRP was coded in C++, while all the other algorithms in this paper were coded in Python 3.0. 

The ratio of the training set to the test set size is 8:2 for all datasets.
All the stacking methods tested were based on the same six base models: random forests (RF), support vector machine (SVM), 
$k$-nearest neighbors ($k$NN), logistic regression (LR), XGBoost (XGB), and feedforward neural networks (FNN).
XGB and FNN were implemented by using Python packages XGBoost and PyTorch, respectively, and all the others were implemented by using scikit-learn (sklearn) package.
For SVM, we used C-support vector classification (SVC) with kernel set as `linear,' and all the learning algorithms of all the other base models were set to their default parameters.
All original features from the datasets were used directly (no hyperparameter tuning or feature engineering was applied).

For the meta-model, if a learning algorithm used in a base model was reused, it was implemented in the same manner as the base model. The time limit for solving any MWRP instance with ILS was set to 60 seconds per run, while no time limit was imposed on other algorithms.
For all stacking approaches utilizing different meta-models, out-of-fold predictions were consistently applied, with the number of folds set to six for all algorithms across all datasets.

For each dataset, a single run of each algorithm was executed, with the random seed consistently set to the same value 1 across all algorithms and datasets to ensure reproducibility.

\subsection{Computational Results}
We use the accuracy (ratio of corrected predictions to all predictions) on the test dataset as the evaluation metric, which is the most commonly used evaluation metric in binary classification.

First, we compare the proposed methods with the learning methods used in the base models without stacking. \tablename~\ref{tbl:accuracy-single} shows the results, with accuracies on the test datasets presented as percentages.

\begin{table*}[ht]
\centering \fontsize{9pt}{11pt}\selectfont
    \caption{Comparison results in percentage accuracy with other stacking approaches based on binary predictions.}\label{tbl:accuracy-bin}
    \begin{tabular}{lrrrrrrrrrrrr}
    \hline
    && \multicolumn{6}{c}{Stacking using binary predictions} && \multicolumn{3}{c}{Stacking with MWRP}                            \\ \cline{3-8} \cline{10-12} 
    dataset && \multicolumn{1}{c}{RF} & \multicolumn{1}{c}{SVM} & \multicolumn{1}{c}{$k$NN} & \multicolumn{1}{c}{LR} & \multicolumn{1}{c}{XGB} & \multicolumn{1}{c}{FNN} && \multicolumn{1}{c}{0-MWRP} & \multicolumn{1}{c}{1-MWRP} & \multicolumn{1}{c}{mix-MWRP}\\ \hline
Income	&&	86.81\%	&	86.84\%	&	84.47\%	&	85.86\%	&	86.79\%	&	86.89\%	&&	86.46\%	&	86.41\%	&	86.44\%	\\
Bank	&&	91.61\%	&	91.43\%	&	90.77\%	&	91.15\%	&	91.53\%	&	91.62\%	&&	91.36\%	&	91.78\%	&	91.39\%	\\
LFW	&&	61.82\%	&	62.50\%	&	62.73\%	&	62.95\%	&	61.82\%	&	62.73\%	&&	62.27\%	&	63.18\%	&	64.09\%	\\
Banknote	&&	99.64\%	&	99.64\%	&	99.64\%	&	99.64\%	&	99.64\%	&	99.64\%	&&	99.64\%	&	99.64\%	&	100.00\%	\\
Credit	&&	75.50\%	&	77.00\%	&	76.00\%	&	77.50\%	&	76.50\%	&	77.50\%	&&	77.00\%	&	79.50\%	&	77.50\%	\\
Cancer	&&	97.37\%	&	97.37\%	&	96.49\%	&	95.61\%	&	96.49\%	&	95.61\%	&&	96.49\%	&	97.37\%	&	98.25\%	\\
Ionosphere	&&	92.96\%	&	92.96\%	&	91.55\%	&	94.37\%	&	91.55\%	&	81.69\%	&&	90.14\%	&	91.55\%	&	92.96\%	\\
Sonar	&&	76.19\%	&	69.05\%	&	76.19\%	&	71.43\%	&	76.19\%	&	76.19\%	&&	83.33\%	&	90.48\%	&	90.48\%	\\\hline
Average	&&	85.24\%	&	84.60\%	&	84.73\%	&	84.81\%	&	85.06\%	&	83.98\%	&&	85.84\%	&	87.49\%	&	87.64\%	\\\hline
    \end{tabular}
\end{table*}

\begin{table*}[ht]
\centering \fontsize{8.5pt}{11pt}\selectfont
    \caption{Comparison results in percentage accuracy with other stacking approaches based on probability predictions.}\label{tbl:accuracy-prob}
    \begin{tabular}{lrrrrrrrrrrrr}
    \hline
    && \multicolumn{6}{c}{Stacking using probability predictions} && \multicolumn{3}{c}{Stacking with MWRP}                            \\ \cline{3-8} \cline{10-12} 
    dataset && \multicolumn{1}{c}{RF} & \multicolumn{1}{c}{SVM} & \multicolumn{1}{c}{$k$NN} & \multicolumn{1}{c}{LR} & \multicolumn{1}{c}{XGB} & \multicolumn{1}{c}{FNN} && \multicolumn{1}{c}{0-MWRP} & \multicolumn{1}{c}{1-MWRP} & \multicolumn{1}{c}{mix-MWRP}\\ \hline
Income	&&	86.09\%	&	86.82\%	&	84.82\%	&	86.76\%	&	86.13\%	&	86.97\%	&&	86.46\%	&	86.41\%	&	86.44\%	\\
Bank	&&	91.21\%	&	91.89\%	&	90.53\%	&	91.65\%	&	91.25\%	&	91.81\%	&&	91.36\%	&	91.78\%	&	91.39\%	\\
LFW	&&	60.45\%	&	65.00\%	&	57.73\%	&	64.77\%	&	58.86\%	&	65.68\%	&&	62.27\%	&	63.18\%	&	64.09\%	\\
Banknote	&&	100.00\%	&	100.00\%	&	100.00\%	&	100.00\%	&	100.00\%	&	99.64\%	&&	99.64\%	&	99.64\%	&	100.00\%	\\
Credit	&&	74.00\%	&	75.00\%	&	74.50\%	&	76.00\%	&	76.00\%	&	76.00\%	&&	77.00\%	&	79.50\%	&	77.50\%	\\
Cancer	&&	98.25\%	&	96.49\%	&	96.49\%	&	97.37\%	&	97.37\%	&	95.61\%	&&	96.49\%	&	97.37\%	&	98.25\%	\\
Ionosphere	&&	91.55\%	&	91.55\%	&	95.77\%	&	91.55\%	&	92.96\%	&	87.32\%	&&	90.14\%	&	91.55\%	&	92.96\%	\\
Sonar	&&	85.71\%	&	78.57\%	&	69.05\%	&	78.57\%	&	78.57\%	&	78.57\%	&&	83.33\%	&	90.48\%	&	90.48\%	\\\hline
Average	&&	85.91\%	&	85.67\%	&	83.61\%	&	85.83\%	&	85.14\%	&	85.20\%	&&	85.84\%	&	87.49\%	&	87.64\%	\\\hline
    \end{tabular}
\end{table*}

The results in \tablename~\ref{tbl:accuracy-single} indicate that the MWRP-based stacking methods generally outperform or match the accuracy of individual base models across various datasets. The average accuracy of all the MWRP-based methods exceeds that of any single base model, with notable improvements on datasets like Credit, Cancer and Ionosphere.
The stacking methods leverage the strengths of different models, leading to enhanced performance, particularly with the MWRP-based method, which achieves the highest overall accuracy. This demonstrates the effectiveness of MWRP-based stacking in enhancing predictive accuracy compared to using individual base models alone.

Next, we compare stacking methods that utilize different meta-models on the test datasets.
The results in Tables~\ref{tbl:accuracy-bin} and \ref{tbl:accuracy-prob} compare the performance of MWRP-based stacking methods with other stacking approaches using binary and probability predictions from base models, respectively. Across both tables, the MWRP-based methods (0-MWRP and 1-MWRP, and mix-MWRP) consistently achieve competitive or superior average accuracies. Specifically, the mix-MWRP method exhibits the highest average accuracy in both comparisons, suggesting that it effectively leverages the strengths of base models. Notably, the MWRP-based stacking methods outperform other approaches on key datasets such as Credit, Cancer and Ionosphere, demonstrating their robustness and efficiency in improving predictive accuracy across different prediction formats.

However, for datasets with a large number of samples (Income and Bank), the MWRP-based stacking methods are not clear winners compared to other stacking approaches using probability predictions. 
This suggests that the solutions to the MWRP provided by ILS may still need improvement due to the 60-second time limit.

\section{Possible Extensions}

The MWRP-based technique presented in this paper demonstrates strong potential for various extensions beyond the scenarios tested.
We believe that this technique can be extended to several other areas, including feature engineering and hyperparameter tuning of base models.
By introducing base models with the same learning algorithm but different features or different parameters, the MWRP-based meta-model can obtain a solution that not only indicates which selections are good but also quantifies their effectiveness.
Additionally, the approach discussed in this paper can also be extended to multi-class classification by existing approaches such as one-vs-all (OvA) and one-vs-one (OvO).
In this section, we discuss two specific extensions of the MWRP-based technique: its application as an interpretable binary classifier and its ability to handle imbalanced datasets.

\subsection{Interpretable Binary Classifier}

One key extension of the MWRP-based technique is its direct application as a binary classifier. 
This approach leverages on the inherent interpretability of the MWRP model, making it an ideal tool for decision-makers who need models that are both accurate and easy to understand.

To apply the MWRP as a binary classifier, it is necessary to ensure that all input data is numeric. Categorical data should be transformed or encoded appropriately before applying the MWRP-based classifier.

Compared to traditional classifiers, the MWRP-based classifier offers the dual benefits of interpretability and operational simplicity. This makes it particularly valuable in fields where transparency is crucial, such as healthcare, finance, or legal decision-making. The classifier clearly indicates which features are selected and the extent of their contribution to the final decision, providing valuable insights into the rationale of the model.

\subsection{Dealing with Imbalanced Datasets}

Another promising extension of the MWRP-based technique is its application to imbalanced datasets. Imbalanced datasets, where one class significantly outnumbers the other, pose a challenge for many machine learning models, often leading to biased predictions that favor the majority class.
Because the MWRP-based approach can natively set weights by modifying \eqref{0mwr} and \eqref{1mwr} for each sample, it offers a natural solution to this issue by easily incorporating different weights among samples, thereby addressing the imbalance.
By carefully weighting samples, the MWRP model can improve its sensitivity to the minority class, leading to better overall performance metrics such as recall, F1-score, and AUC.

Compared to traditional techniques for handling imbalanced datasets, such as SMOTE (synthetic minority over-sampling technique) or class weighting in models like SVMs or neural networks, the MWRP-based approach offers a unified and interpretable framework that can seamlessly integrate with the stacking ensemble learning process. This makes it a powerful tool for addressing class imbalance while maintaining the overall integrity and interpretability of the model.

\section{Conclusion}
In this paper, we presented a novel stacking ensemble learning framework that utilizes the Maximum Weighted Rectangle Problem (MWRP) as a core component for constructing meta-models. Our approach leverages the strengths of various base models while maintaining high interpretability and low computational complexity. Through extensive experiments, we demonstrated that the MWRP-based stacking methods generally outperform or match the accuracy of individual base models and other traditional stacking approaches across multiple datasets.

While the MWRP-based techniques showed promising results, particularly with the 1-MWRP method achieving the highest overall accuracy, challenges remain for datasets with large numbers of samples, where the 60-second time limit imposed on solving the MWRP may lead to suboptimal solutions. Despite these challenges, the proposed framework offers significant advantages in terms of transparency and ease of operation, making it a valuable tool in scenarios where model interpretability is critical.

Additionally, we identified potential extensions of the MWRP-based technique, including its application as a binary classifier and its ability to handle imbalanced datasets by adjusting the weights of individual samples. These extensions highlight the versatility and broad applicability of the MWRP-based approach, opening avenues for further research and development.

Overall, the MWRP-based stacking method represents a powerful and flexible tool for improving predictive performance in binary classification tasks, with the potential for future adaptations and enhancements to address a wider range of machine learning challenges.

% Acknowledgments here
\section*{Acknowledgement}
% Enter the text of acknowledgments here
This work was supported by JSPS KAKENHI [Grant No.19K11843].

% \bibliographystyle{unsrt}
% \bibliography{references}

\end{document}